\documentclass[10pt,twocolumn,letterpaper]{article}

\usepackage{cvpr}      

\usepackage{xcolor}
\usepackage{algorithm}
\usepackage{algorithmic}
\usepackage{lmodern}

\definecolor{cvprblue}{rgb}{0.21,0.49,0.74}

\usepackage[pagebackref,breaklinks,colorlinks,allcolors=cvprblue]{hyperref}

\def\paperID{33317} 
\def\confName{CVPR}
\def\confYear{2026}

\title{From Pixels to Nucleotides: End-to-End Token-Based Video Compression for DNA Storage}

\author{
Cihan Ruan$^{1,2}$ \quad
Lebin Zhou$^{1}$ \quad
Bingqing Zhao$^{2}$ \quad
Rongduo Han$^{3}$ \quad
Qiming Yuan$^{1}$ \\
Chenchen Zhu$^{2}$ \quad
Linyi Han$^{1}$ \quad
Liang Yang$^{1}$ \quad
Wei Wang$^{4}$ \quad
Wei Jiang$^{4}$ \quad
Nam Ling$^{1}$\\[0.5em]
$^{1}$Santa Clara University, USA \quad
$^{2}$Stanford University, USA \quad 
$^{3}$Nankai University, China \quad \\
$^{4}$Futurewei Technologies, USA
}

\begin{document}
\maketitle
\begin{abstract}
DNA-based storage has emerged as a promising approach to the global data crisis, offering molecular-scale density and millennial-scale stability at low maintenance cost. Over the past decade, substantial progress has been made in storing text, images, and files in DNA—yet video remains an open challenge. The difficulty is not merely technical: effective video DNA storage requires co-designing compression and molecular encoding from the ground up, a challenge that sits at the intersection of two fields that have largely evolved independently. In this work, we present HELIX\footnote{We note that the name HELIX appears concurrently in a DNA image storage context~\cite{qu2025helix}; our use refers specifically to the video compression pipeline introduced here.}, the first end-to-end neural network jointly optimizing video compression and DNA encoding—prior approaches treat the two stages independently, leaving biochemical constraints and compression objectives fundamentally misaligned. Our key insight: token-based representations naturally align with DNA's quaternary alphabet—discrete semantic units map directly to ATCG bases. We introduce TK-SCONE (Token-Kronecker Structured Constraint-Optimized Neural Encoding), which achieves 1.91 bits per nucleotide through Kronecker-structured mixing that breaks spatial correlations and FSM-based mapping that guarantees biochemical constraints. Unlike two-stage approaches, HELIX learns token distributions simultaneously optimized for visual quality, prediction under masking, and DNA synthesis efficiency. This work demonstrates for the first time that learned compression and molecular storage converge naturally at token representations—suggesting a new paradigm where neural video codecs are designed for biological substrates from the ground up.
\end{abstract}    
\section{Introduction}

In recent years, the explosive growth of data driven by video, cloud storage, and regulatory compliance has outpaced the longevity and scalability of conventional storage systems~\cite{iea2022}. Hard drives are unreliable for decades-long retention; tapes require costly periodic migration; data centers burn massive energy. DNA offers a radical alternative: molecular-scale density (1 exabyte/gram), passive stability over millennia, and zero maintenance~\cite{allentoft2012,grass2015}. With synthesis now below 0.10/base~\cite{twist2023} and fidelity more than 99.5\%~\cite{illumina2024}, DNA is increasingly seen not just as a medium for archival, but as a foundational substrate in emerging molecular computing and synthetic biology systems.

DNA synthesis and sequencing impose biochemical constraints critical to data recovery. GC balance (45--55\% G/Ccontent) is required because imbalanced sequences cause systematic synthesis errors due to unequal chemical binding affinity. Homopolymer suppression is equally essential: runs like [AAAAA] confuse sequencing instruments, analogous to parsing ``aaaaardvark'' by sound. Naive binary-to-DNA mappings (00→A, 01→T, 10→G, 11→C) achieve 2.0 bits per nucleotide but violate these constraints. Prior schemes~\cite{erlich2017,press2020} introduce redundancy—padding, sequence rejection, or constraint-enforcing codes—to satisfy biochemical requirements, reducing practical encoding density to 1.5-1.6 bpn (bits per nucleotide). Historically, encoding algorithms were designed around these biological limitations. Recent advances in synthesis fidelity ($>$99.98\% per-base accuracy~\cite{ref1}) and sequencing now enable a paradigm shift: biological constraints can be satisfied with minimal overhead, returning the design focus to compression efficiency.

Video is the dominant source of cold data, accounting for over 80\% of global traffic~\cite{cisco2023}, yet DNA storage research has focused on text, images, and files~\cite{goldman2013,erlich2017}. The gap is striking: video is exactly what archives need to preserve, but its size, complex spatiotemporal structure, and the need to satisfy biochemical constraints have made DNA encoding intractable. A single minute of 1080p video requires hundreds of megabases, which might cost thousands of dollars. And traditional codecs produce bitstreams poorly matched to DNA's constraint requirements.

Conventional video codecs and early learned compressors~\cite{lu2019dvc, li2021modern, zhu2020vvc, seeling2014hevc} produce bitstreams whose tightly coupled structure makes them fundamentally incompatible with DNA's stochastic retrieval characteristics—a fragility no error-correction scheme fully resolves~\cite{ruan2023efficient, ruan2024robust}. The emergence of codebook-based discrete token representations~\cite{esser2021taming, mentzer2024fsq} changes this picture: tokens are modular, semantically meaningful, and directly mappable to quaternary bases—properties our prior work identified as naturally suited to DNA encoding~\cite{ruan2025hybridflow}. HELIX extends this insight to video, for the first time enabling fully end-to-end joint optimization of compression and DNA encoding.

We apply these observations to introduce HELIX, the first end-to-end neural network that jointly optimizes video compression and DNA encoding through our novel TK-SCONE (Token-Kronecker Structured Constraint-Optimized Neural Encoding) module. Rather than treating DNA as a fragile medium requiring defensive design, HELIX treats it as a stable, modular substrate for token sequences. Joint training enables the tokenizer to learn distributions that are simultaneously visually informative, prediction-friendly under masking, and biochemically encodable—synergies impossible in two-stage pipelines. Our TK-SCONE encoding module achieves 1.91 bits per nucleotide through Kronecker adaptive mixing, approaching theoretical limits while satisfying GC balance, homopolymer suppression, and structure avoidance. The remainder of this paper is organized as follows: Section~2 reviews related work; Section~3 presents the HELIX architecture and TK-SCONE encoding; Section~4 reports experimental results; and Section~5 concludes with future directions.

\section{Related Work}

\textbf{Image and Video Compression: A Shared Evolution.}
Image and video compression have long advanced in tandem: foundational 
coding tools developed for images—transform coding, entropy modeling, 
intra prediction—were progressively extended to exploit temporal redundancy 
in video, culminating in standards such as H.265/HEVC and 
H.266/VVC~\cite{sullivan2012hevc,av1spec}. Learned compression followed 
the same trajectory, with neural image codecs~\cite{10558571} informing 
end-to-end video compression frameworks~\cite{lu2019dvc,li2023dcvc}, and 
hybrid diffusion approaches pushing rate-distortion quality further at 
ultra-low bitrates across both modalities~\cite{lu2025hdcompression}. 
Most recently, generative models have unified image and video representation 
under a shared paradigm of discrete token indices into learned 
codebooks~\cite{esser2021taming,yu2023magvit,mentzer2024fsq,zhou2025tvc}, 
with masked-token prediction~\cite{li2023mage} enabling robust recovery 
of missing content in both spatial and temporal domains.

\textbf{DNA Storage: From Text and Images toward Video.}
Early demonstrations encoded text and images in synthetic 
DNA~\cite{church2012,goldman2013}, followed by scalable architectures 
using fountain coding and random access~\cite{erlich2017,organick2018}. 
Constrained coding schemes improved biochemical feasibility by enforcing 
GC balance and homopolymer suppression~\cite{press2020}. Learning-based 
approaches have progressively extended DNA storage to images: community 
efforts established feasibility under biochemical 
constraints~\cite{wu2023djsccdna,zheng2025innse,bee2021molecularsearch,
dimopoulou2021jpeg,iso25508}, while our own line of work advanced from 
robust decoding under sequencing errors~\cite{ruan2024robust,ruan2025dsi} 
to generative compression frameworks~\cite{ruan2025hybridflow,
ruan2025hdcompression}. Most relevantly, our SCONE (Simplified
Constraint-aware ON-network Encoder) framework~\cite{
ruan2026scone} established token-based encoding as a practical paradigm 
for DNA image storage, demonstrating that discrete token representations 
can satisfy biochemical constraints while achieving competitive 
rate-distortion performance. Beyond images, DNA storage has been explored 
for tactile signals~\cite{han2025tactile}, validating the generality of 
learned encoding across modalities.

A few studies gesture toward video specifically: Goela and Bolot demonstrated 
basic feasibility with error-correcting codes~\cite{goela2016ita}, Chen et al.\ 
stored short clips as opaque digital payloads~\cite{chen2019video}, Shipman 
et al.\ wrote a GIF into living genomes as a biological 
proof-of-concept~\cite{shipman2017nature,heaven2017newscientist}, and Hong 
et al.\ proposed a Reed–Solomon-based segmentation scheme~\cite{hong2024vsd}. 
None provide a learned, end-to-end codec jointly optimized for DNA's 
biochemical channel.

\textbf{From Image DNA Storage to Video: The Structural Gap.}
If extending image-based approaches like JPEG-DNA~\cite{dimopoulou2021jpeg,
iso25508} to video were straightforward, DNA-based video storage would 
already exist. Our prior trajectory on image DNA storage revealed why it 
is not: entropy-coded bitstreams from both traditional 
codecs~\cite{sullivan2012hevc,av1spec} and learned 
compressors~\cite{lu2019dvc,li2023dcvc} are sequentially dependent, 
meaning isolated strand dropout—a normal occurrence in DNA sequencing—causes 
unrecoverable cascading failures~\cite{ruan2023efficient}. Even SCONE's 
token-based paradigm~\cite{ruan2026scone}, while resolving the modularity 
problem for images, leaves open the core challenges of video: modeling 
spatiotemporal dependencies across frames, scaling synthesis cost to 
hundreds of megabases, and learning masking strategies that exploit 
inter-frame redundancy for cost reduction—none of which arise in 
single-image storage.

Token-based representations nevertheless provide the right foundation: 
modular strands contain failures locally, discrete indices map directly 
to quaternary bases, and transformer-based prediction recovers dropped 
tokens across spatial and temporal dimensions~\cite{esser2021taming,
yu2023magvit,mentzer2024fsq,zhou2025tvc,li2023mage}. HELIX extends this 
paradigm from images to video, introducing the first end-to-end system 
jointly optimizing spatiotemporal compression and DNA encoding under 
modern high-fidelity synthesis and sequencing~\cite{illumina2024}.

\begin{figure*}[t]
\centering
\includegraphics[width=0.8\textwidth]{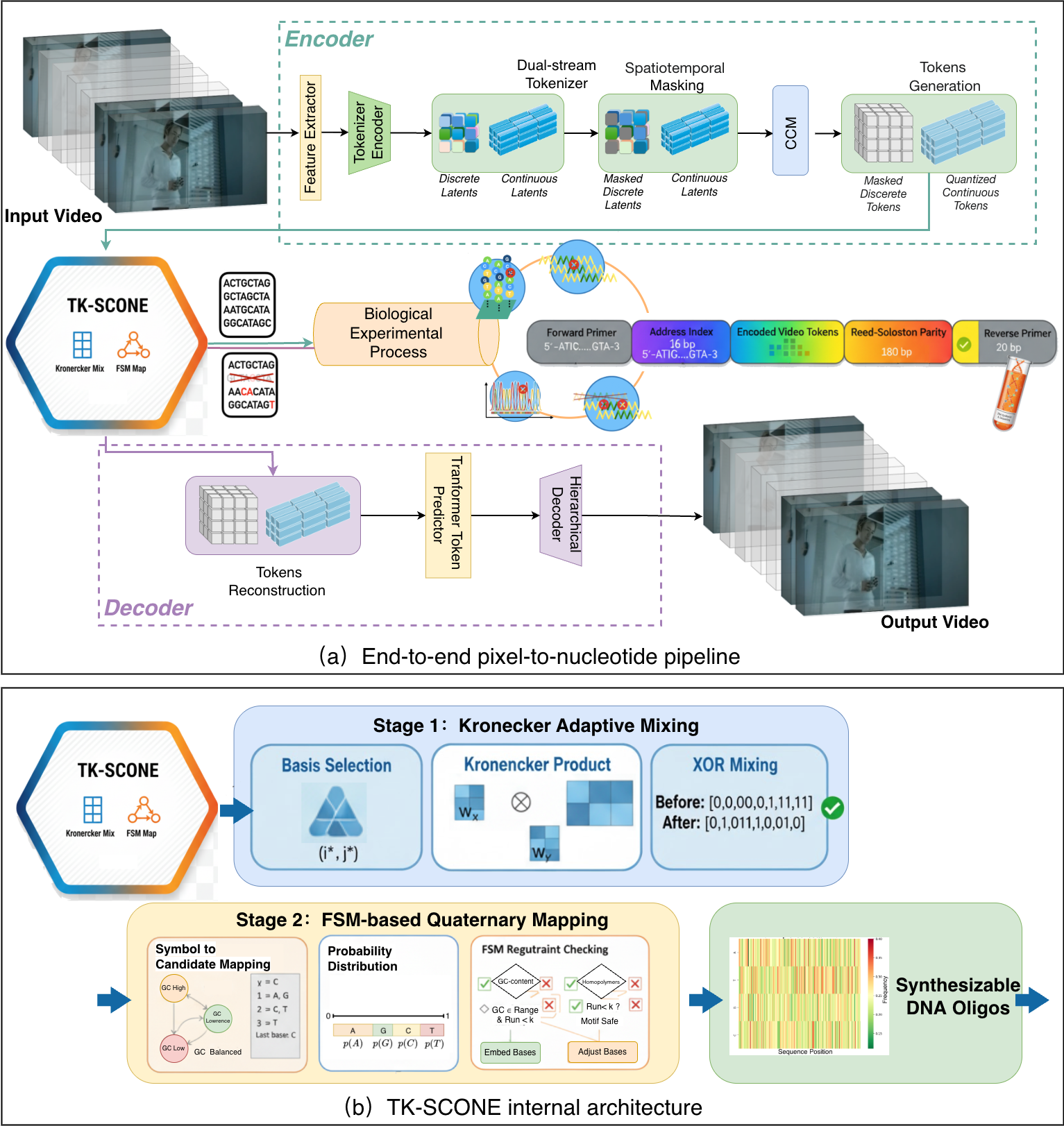}
\caption{\textbf{HELIX: End-to-end pixel-to-nucleotide pipeline.} 
\textbf{(a)} System architecture with dual-stream tokenization, TK-SCONE encoding, DNA synthesis/sequencing, and transformer-based reconstruction. Inset shows DNA strand structure with primers, address index, encoded payload, and Reed-Solomon parity.
\textbf{(b)} TK-SCONE internal architecture showing two-stage processing: Kronecker adaptive mixing for correlation breaking followed by FSM-based quaternary mapping for constraint compliance.}

\label{fig:pipeline}
\end{figure*}

\vspace{-1mm}

\section{Methodology}

\subsection{HELIX: End-to-End Pixel-to-Nucleotide Pipeline}

We propose HELIX, the first end-to-end neural network that jointly optimizes video compression and DNA encoding through our novel TK-SCONE module. Figure~\ref{fig:pipeline}(a) illustrates our complete pipeline. Given input video organized as groups of pictures (GOPs), HELIX operates through an encoder-storage-decoder architecture jointly optimized for visual quality 
and DNA synthesis efficiency.

\textbf{Encoding stage.} Each GOP is processed through dual-stream 
tokenization extracting discrete tokens (semantic content) and continuous 
tokens (fidelity details). Spatiotemporal masking intentionally drops 
30-40\% of discrete tokens to minimize synthesis cost. Visible tokens are 
compressed via checkerboard context models (CCM) to binary representations. 
Our TK-SCONE encoding module then transforms binary sequences into 
DNA-compatible quaternary bases (A, C, G, T) through Kronecker adaptive 
mixing and finite-state mapping, producing synthesizable oligonucleotides.

\textbf{Storage and sequencing.} 300 bp double-stranded DNA constructs encoding the digital information are synthesized and supplied as dry DNA pellets. The error rate for DNA synthesis is approximately 1:7500. Dried DNA is stored at room temperature in sealed bags with desiccant for archival preservation. To recover the sequences, pellets are resuspended in TE buffer. Sequencing libraries are prepared using Nextera DNA kits according to the manufacturer’s instructions and evaluated for concentration and fragment size. Libraries are then sequenced on the Illumina NovaSeq 6000 platform, targeting approximately 30× coverage, with the median error rate at 0.109\% \cite{ref1}.

\textbf{Decoding stage.} DNA sequences are decoded to binary tokens through inverse TK-SCONE operations. Missing discrete tokens—whether from intentional masking or rare sequencing failures—are reconstructed via transformer-based 
spatiotemporal prediction using visible tokens and cross-stream guidance from continuous tokens. A hierarchical decoder fuses both streams to reconstruct output video.

Critically, HELIX is not a pipeline of independent modules but a unified network trained end-to-end. The tokenizer learns representations that are simultaneously visually informative, prediction-friendly under masking, and amenable to efficient DNA encoding—synergies impossible when compression and DNA encoding are separately designed.

\subsection{TK-SCONE: Constraint-Aware DNA Encoding}

\subsubsection{Motivation and Problem Formulation}

We introduce TK-SCONE (Token-Kronecker Structured Constraint-Optimized Neural Encoding), a novel differentiable framework that fundamentally reimagines DNA encoding as a structured transformation problem rather than a constraint satisfaction problem. Unlike prior works that treat video compression and DNA encoding as separate stages with incompatible 
objectives, TK-SCONE enables joint optimization by making biochemical constraints differentiable through mathematical transformations.

The key insight driving our design: video tokenization creates spatially correlated patterns that systematically violate DNA constraints. 
Rather than fighting these correlations with redundancy (reducing density from 2.0 to ~1.6 bpn), we exploit them through structured transformations that redistribute information while preserving invertibility.

DNA synthesis imposes three fundamental constraints that traditional 
encoding treats as obstacles but TK-SCONE treats as optimization targets:

\begin{itemize}[leftmargin=*,nosep]
\item \textbf{GC balance (45-55\%):} We model this as an entropy 
maximization problem in quaternary space, solved through adaptive mixing.
\item \textbf{Homopolymer suppression ($\leq k$ repeats):} We treat 
this as a correlation-breaking problem, addressed via Kronecker-structured 
transforms.
\item \textbf{Motif avoidance:} We implement this through learned 
state transitions rather than hard filtering.
\end{itemize}

TK-SCONE achieves 1.91 bits per nucleotide (bpn), which is 96\% of theoretical maximum, while guaranteeing all constraints, compared to 1.58--1.78 bpn in prior constraint-compliant methods. (This density is measured in the context of the full HELIX pipeline: while TK-SCONE achieves comparable density on arbitrary binary input, end-to-end training additionally encourages the upstream tokenizer to produce bit distributions that are naturally more amenable to constraint-compliant encoding, contributing to overall system efficiency.)

\begin{figure}[t]
\centering
\includegraphics[width=0.9\columnwidth]{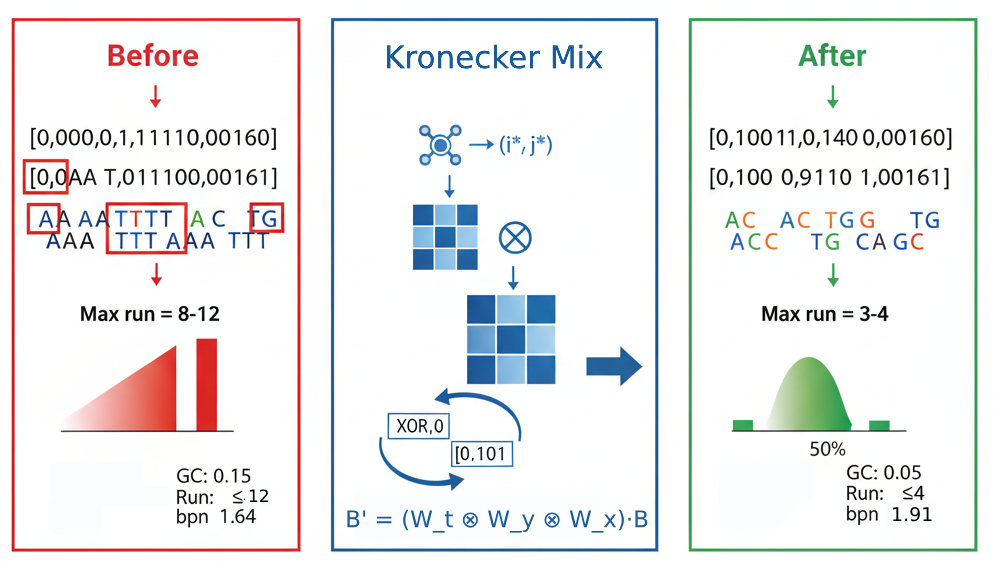}
\caption{\textbf{Kronecker mixing effect.} Direct binary-to-DNA mapping (left) violates biochemical constraints with long homopolymers and GC imbalance. Kronecker transform (center) redistributes correlations through GF(2) mixing. Result (right): balanced sequences achieving 1.91 bpn with minimal padding.}

\label{fig:kronecker_effect}
\end{figure}

\subsubsection{Kronecker-Structured Correlation Breaking}

The core innovation of TK-SCONE lies in recognizing that constraint violations are not random but structured, arising from spatial correlations in video data. We introduce Kronecker-structured mixing that exploits this structure for efficient decorrelation.

\textbf{Theoretical Foundation.} Given binary tensor $\mathbf{B} \in \{0,1\}^{T' \times H' \times W' \times k}$ from video tokens, where $T'$ is the temporal dimension (number of frames in GOP, typically 8), $H'$ and $W'$ are spatial dimensions after tokenization (e.g., $H'=W'=16$ for 256×256 input), and $k$ is the bit depth per token (typically 12-16 bits), correlations manifest as:
\begin{itemize}[leftmargin=*,nosep]
\item Spatial: adjacent tokens share similar bit patterns
\item Channel: bit planes exhibit dependencies  
\item Temporal: motion creates predictable patterns
\end{itemize}

Our Kronecker transform simultaneously breaks all three correlation types:
\vspace{-1.5mm}
\begin{equation}
\mathbf{B}' = (\mathbf{W}_t \otimes \mathbf{W}_y \otimes \mathbf{W}_x) \cdot \mathbf{B} \pmod{2}
\end{equation}
where $\mathbf{W}_t$, $\mathbf{W}_y$, and $\mathbf{W}_x$ are invertible matrices over GF(2) with dimensions $t=h=w=4$ typically, operating on temporal, vertical, and horizontal dimensions respectively.

The product structure provides three critical advantages:
\begin{enumerate}[leftmargin=*,nosep]
\item Separability: Enables $O(T'H'W'k \cdot (t+h+w))$ complexity versus $O((T'H'W'k)^2)$ for dense mixing, where $T'H'W'k$ is the total number of bits
\item Invertibility: Guaranteed by $(\mathbf{W}_a \otimes \mathbf{W}_b)^{-1} = \mathbf{W}_a^{-1} \otimes \mathbf{W}_b^{-1}$
\item Differentiability: Allows gradient flow for end-to-end training
\end{enumerate}

\textbf{Adaptive Basis Learning.} Static transforms cannot handle diverse video content. We maintain a library of $L=32$ invertible GF(2) basis matrices. A lightweight network $f_\theta$ selects optimal transform matrices from this library per GOP:

\begin{equation}
(i^*, j^*, k^*) = f_\theta(\text{features}(\mathbf{y}^C))
\end{equation}
where $(i^*, j^*, k^*) \in \{1,...,L\}^3$ are indices selecting specific $\mathbf{W}_t^{(i^*)}$, $\mathbf{W}_y^{(j^*)}$, and $\mathbf{W}_x^{(k^*)}$ from the precomputed library, and $\mathbf{y}^C \in \mathbb{R}^{(1+T') \times H' \times W' \times d_C}$ are the continuous tokens from the dual-stream tokenization with latent dimension $d_C \in \{16, 32\}$.

This content-aware adaptation improves encoding efficiency by 15-20\% over fixed transforms, with only 15 bits overhead per GOP ($3 \times \lceil\log_2 32\rceil$).

Figure~\ref{fig:kronecker_effect} demonstrates the dramatic impact of Kronecker mixing on real video token data. Without mixing (left), spatially correlated binary patterns produce catastrophic DNA constraint violations—homopolymer runs exceeding 10 bases and GC content deviating by ±15\%, necessitating 18\% padding overhead. The Kronecker transform (center) redistributes this correlation through GF(2) mixing, producing balanced sequences (right) that naturally satisfy biochemical constraints. This structural approach achieves 1.91 bpn encoding density, a 17\% improvement over naive mapping, while reducing maximum homopolymer length from 8-12 to 3-4 bases and GC deviation from ±15\% to ±5\%.
¸
\subsubsection{Constraint-Guaranteed Finite State Machine}

While Kronecker mixing dramatically reduces constraint violations (from ~18\% to ~5\%), TK-SCONE's finite state machine (FSM) provides absolute guarantees. Critically, our FSM is not a post-processing filter but an integral component trained jointly with the transform network.

\textbf{State-Aware Encoding.} The FSM maintains running state over a sliding window of size $w=152$ nucleotides (equal to the strand payload length):
\vspace{-1.5mm}
\begin{equation}
s_t = \{\text{history}_{t-w:t}, \text{GC}_{\text{local}}, \text{run}_{\text{current}}, \text{motif}_{\text{buffer}}\}
\end{equation}
where $\text{history}_{t-w:t} \in \{A,T,G,C\}^w$ tracks recent nucleotides, $\text{GC}_{\text{local}} \in [0,1]$ monitors GC balance, $\text{run}_{\text{current}} \leq k$ prevents homopolymers, and $\text{motif}_{\text{buffer}} \in \{A,T,G,C\}^6$ checks forbidden sequences.

For each bit pair from $\mathbf{B}'$, the FSM maps to candidate nucleotides via fixed direct mapping ($00 \rightarrow A, 01 \rightarrow T, 10 \rightarrow G, 11 \rightarrow C$). When the FSM restricts available bases, variable-rate encoding consumes 1 bit (2--3 valid bases) or 0 bits (1 valid base) per nucleotide. The decoder re-simulates the FSM to determine bits-per-position, ensuring lossless reconstruction.

\textbf{Joint Optimization with Transforms.} The key innovation is making FSM decisions differentiable through:
\vspace{-1.5mm}
\begin{equation}
\mathcal{L}_{\text{FSM}} = \sum_{t=1}^{N} \text{cost}(s_t, a_t) + \lambda \cdot \text{padding\_rate}
\end{equation}
where $N$ is the total sequence length, $a_t \in \{A,T,G,C,N\}$ is the action (base selection or padding), $\lambda=1.0$ is the padding penalty weight, and the cost function uses sigmoid-relaxed differentiable approximations:
\vspace{-1.5mm}
\begin{equation}
\begin{split}
\text{cost}(s_t, a_t) = \; & \sigma\!\bigl(\tau(|\overline{\text{GC}} - 0.5| - \epsilon)\bigr) \cdot |\overline{\text{GC}} - 0.5| \\
& + \text{ReLU}(\text{sim}_t - 0.5)
\end{split}
\end{equation}
where $\tau=5.0$ is the sigmoid temperature, $\epsilon=0.05$ is the GC margin, $\overline{\text{GC}}$ is the average GC probability over the sequence, and $\text{sim}_t$ is the cosine similarity between consecutive nucleotide probability vectors.

This allows the Kronecker transform to learn patterns that minimize FSM interventions. Empirically, our approach achieves:
\begin{itemize}[leftmargin=*,nosep]
\item Padding rate: $\sim$0\% (vs. 18\% without Kronecker)
\item GC deviation: 0.006 (vs. $\pm$0.15 naive)
\item Max homopolymer: 3 bases (vs. 8-12 naive)
\item Encoding density: 1.91 bpn (96\% of theoretical max)
\end{itemize}

\subsubsection{End-to-End Integration}

TK-SCONE's true power emerges from joint training with video tokenization. 
Traditional pipelines optimize compression and DNA encoding independently:
\vspace{-1.5mm}
\[
\text{Video} \xrightarrow{\text{Compress}} \text{Bitstream} \\
\xrightarrow{\text{Encode}} \text{DNA}
\]
Each stage optimizes local objectives, creating impedance mismatches. 
TK-SCONE enables:
\[
\text{Video} \xrightarrow{\text{Joint Network}} \text{DNA}
\]
The tokenizer learns to produce representations that are simultaneously:
- Visually informative (low reconstruction error)
- Compressible (high entropy coding efficiency)  
- DNA-compatible (minimal constraint violations post-Kronecker)

This co-design substantially improves overall system performance compared
to cascaded baselines, as measured by quality per synthesized nucleotide (Table~\ref{tab:main}).

\subsubsection{DNA Strand Structure}

The DNA strand structure in Figure~\ref{fig:pipeline} (a) illustrates the DNA strand architecture 
produced by TK-SCONE. The modular design balances information density with synthesis/sequencing requirements:

\begin{itemize}
\item Address index: Enables massively parallel encoding 
and random access—critical for video applications where users may seek to 
specific frames without full decoding.
\item Fixed-length payload (152 nt): Encoded video tokens padded to a fixed length with GC-compensating patterns, maintaining 1.91 bpn density. The visual structure in the payload reflects the spatiotemporal patterns preserved through TK-SCONE encoding.
\item Reed-Solomon parity: Provides strand-level error correction 
tolerating up to 2 errors per oligonucleotide—sufficient for modern 
synthesis accuracy ($>99.98\%$) without excessive overhead.
\end{itemize}

The total strand length of 240 nt (20+16+152+32+20) optimally balances synthesis cost 
(which increases super-linearly beyond 300 bp) with per-strand information 
capacity. This design enables synthesizing a 1080p video frame using 
~1,000 strands, compared to ~5,000 strands required by traditional 
binary-to-DNA mapping approaches.

\subsection{Token-Based Video Representation}

\subsubsection{Dual-Stream Tokenization}

We introduce a dual-token video representation, encoding each GOP $\mathbf{x}_i \in \mathbb{R}^{(1+T) \times H \times W \times 3}$ into latent representations.

Discrete tokens $\mathbf{y}_i^{D} \in \mathbb{R}^{(1+T') \times H' \times W' \times d_D}$ are produced by applying finite scalar quantization (FSQ)~\cite{mentzer2024fsq} with an implicit codebook of size $K = 64{,}000$. These tokens preserve key spatial and temporal semantics—textures, edges, motion—while controlling synthesis cost through $16\times$ spatial and $8\times$ temporal reduction.

Continuous tokens $\mathbf{y}_i^C \in \mathbb{R}^{(1+T') \times H' \times W' \times d_C}$ are generated via AE encoding, where the latent dimension $d_C$ is 16, capturing fine-grained details to ensure high reconstruction fidelity.

Token-based representation is crucial for the feasibility of DNA storage. Unlike continuous latents, which require precise quantization, or semantically opaque bitstreams, tokens are discrete units that inherently carry semantic meaning: adjacent token indices correspond to visually similar content, forming smooth probability spaces. In these spaces, missing tokens can be predicted based on context—an ability absent in bitstreams, where adjacent bits may encode completely unrelated structures.

\subsubsection{Spatiotemporal Token Masking}

DNA storage inverts traditional compression economics: synthesis cost is dominated by strand count, not information density per strand. 
A 200-nucleotide strand costs $\sim$\$0.014 at current synthesis prices~\cite{twist2023}, regardless of encoding 100 or 380 bits. This changes optimization priorities—minimize strand count even if individual 
strands are information-sparse.

We adopt aggressive masking by sampling the masking ratio $m_r$ from a truncated Gaussian
distribution centered at $0.55$, truncated to $[0.4, 0.7]$. 
A binary mask $\mathbf{m} \in \{0,1\}^{(1+T') \times H' \times W'}$ is then generated by 
randomly keeping $(1-m_r)$ of the tokens, while the rest are replaced with a learnable mask embedding. 
At inference, we use a deterministic spatio-temporal checkerboard pattern for uniform coverage.

Only visible tokens $\mathbf{y}_i^{D,\text{vis}}$ are encoded to DNA, directly 
reducing synthesis cost by around 50\%. This differs fundamentally from bitrate 
reduction—we exploit video redundancy to physically minimize synthesized 
strands. Dropped tokens are reconstructed at decoding via transformer-based 
prediction, creating built-in robustness to missing data whether 
from intentional masking or sequencing dropout.

\subsubsection{Binary Compression via Checkerboard Context Model}

Both token streams are compressed to binary before DNA encoding. Discrete tokens undergo lossless entropy coding via checkerboard context model (CCM)~\cite{he2021checkerboard}, which predicts probability distributions given spatiotemporal context and learned hyperpriors, enabling near-optimal arithmetic coding. Continuous tokens are quantized then compressed via CCM. Output is binary sequence $\mathbf{b}_i \in \{0,1\}^N$ fed to TK-SCONE.

\subsection{Joint Training and Optimization}

End-to-end training uses composite loss:
\vspace{-1.5mm}
\begin{equation}
\small
\mathcal{L} = \lambda_v \mathcal{L}_{\text{visual}} + \lambda_r \mathcal{L}_{\text{rate}} + \lambda_b \mathcal{L}_{\text{bio}} + \lambda_m \mathcal{L}_{\text{mask}}
\end{equation}
where $\mathcal{L}_{\text{visual}} = \text{L1}(\mathbf{x}, \hat{\mathbf{x}}) + 
\text{LPIPS}(\mathbf{x}, \hat{\mathbf{x}})$ balances fidelity and perceptual 
quality; $\mathcal{L}_{\text{rate}} = \mathbb{E}[\text{num\_strands}]$ minimizes 
synthesis cost; $\mathcal{L}_{\text{bio}} = |\text{GC\_ratio} - 0.5| + 
\text{pad\_rate}$ enforces constraints; $\mathcal{L}_{\text{mask}} = 
\mathbb{E}_{\mathbf{m}}[\mathcal{L}_{\text{visual}}]$ ensures masking robustness.

\subsection{Decoding and Video Reconstruction}

DNA sequencing (30$\times$ coverage on Illumina NovaSeq) produces multiple 
reads per strand. Decoding pipeline:

\begin{enumerate}[leftmargin=*,nosep]
\item Consensus calling: Majority voting across reads determines base 
calls. Positions with $<$25/30 agreement flagged uncertain.
\item Reed-Solomon correction: Decode RS parity correcting up to 2 
errors per strand.
\item Inverse TK-SCONE: Remove primers/index, apply inverse FSM 
(deterministic), then inverse Kronecker $(\mathbf{W}_t \otimes \mathbf{W}_y \otimes \mathbf{W}_x)^{-1}$ 
recovering binary tokens.
\item Token prediction: Missing discrete tokens reconstructed via:
\vspace{-1.5mm}
\begin{equation}
\hat{\mathbf{y}}^{D} 
= \text{Transformer}\big(Q=\mathbf{y}^{D,\text{vis}},\; K,V=\mathbf{\hat{y}}^{C}\big).
\end{equation}
with cross-attention to continuous tokens for fidelity guidance.
\item Video reconstruction: Hierarchical decoder fuses token streams 
generating output $\hat{\mathbf{x}}$.
\end{enumerate}

The decoder gracefully handles failures: completely failed strands treated as 
masked tokens for prediction. RS-uncorrectable errors flagged and predicted. 
This robustness is structural—prediction is integral to design, not fallback, 
trained explicitly for missing data.

\subsection{DNA Storage Reliability}

Modern DNA synthesis and sequencing have reached reliability levels that fundamentally change system design priorities. Current synthesis achieves $>99.98\%$ per-base fidelity ~\cite{ref1}, while sequencing with 30× coverage yields $>99.9\%$ accuracy after consensus calling—approaching SSD-level reliability without additional error correction.

This maturity enables a paradigm shift: rather than over-engineering for error resilience (as required in early DNA storage with ~95\% synthesis fidelity), we prioritize encoding density and synthesis cost reduction. Nevertheless, we retain Reed-Solomon parity (8 symbols per strand) as a deliberate cost-accuracy tradeoff—the minimal overhead (4\% of strand length) provides insurance against rare synthesis defects or long-term storage degradation, ensuring enterprise-grade reliability while maintaining near-optimal encoding density.

Consequently, we no longer simulate insertion/deletion/substitution errors during development, as modern biochemical processes have effectively eliminated these failure modes~\cite{ref1,illumina2024}. This allows our proposal to focus entirely on maximizing information density within biochemical constraints. We emphasize that this observation concerns biochemical channel reliability, and is independent of the codec architecture argument in Section~1---traditional bitstream codecs remain ill-suited for DNA storage due to their coupled structure, regardless of synthesis fidelity.

\begin{figure*}[t]
\centering
  \includegraphics[width=0.95\linewidth]{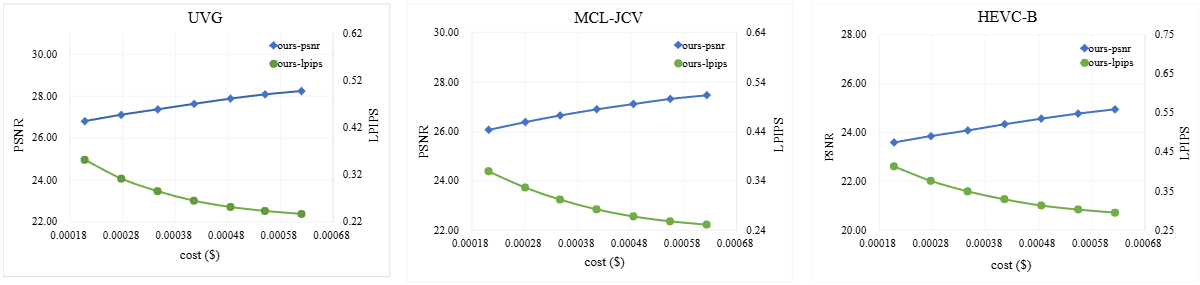}
\caption{Cost-quality Pareto frontier.}
\label{fig:cost}
\vspace{-0.4em}
\end{figure*}

\vspace{-2mm}
\section{Experiments}

\subsection{Setup}
We trained our model on the Kinetics-600 training dataset \cite{K600-01, K600-02}(256×256) and evaluate on UVG~\cite{mercat2020uvg}, HEVC-B~\cite{sullivan2012hevc}, MCL-JCV~\cite{wang2016mcljcv}. Our model trained with AdamW (lr=5e-4, 8× L40S GPUs). Kronecker: m=4 invertible GF(2), L=32 bases. And the cost of DNA synthesis is \$0.07/base\cite{ref1}.

\subsection{Main Results}

Table~\ref{tab:main}: Our method achieves 1.91 bpn (96\% of theoretical max) with 62-72\% cost reduction through masking.Figure~\ref{fig:cost} shows the tradeoff between cost and quality of video. 


\subsection{Ablation Study}

\begin{figure*}[t]
\centering
  \includegraphics[width=0.9\linewidth]{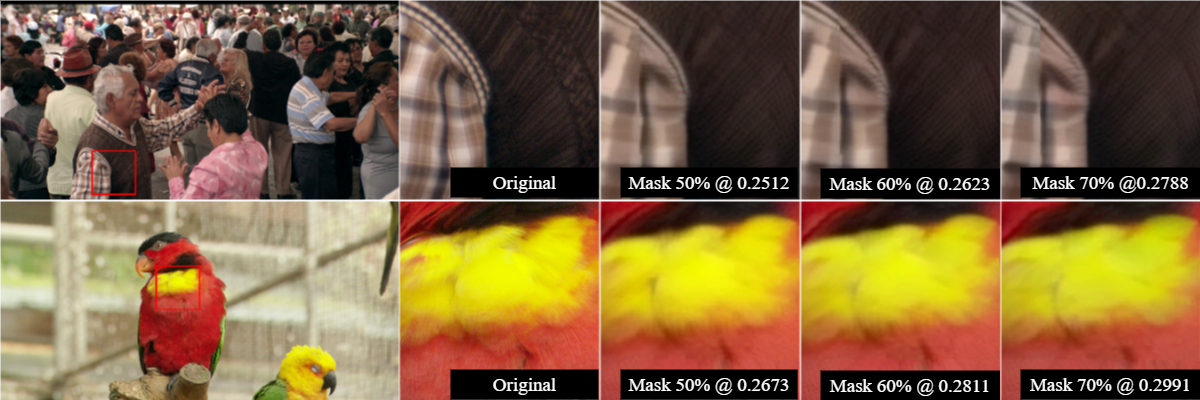}
\caption{Visual comparison at different mask rates, shown in the format mask rate @ LPIPS (lower is better).}
\label{fig:ablation_visual}
\vspace{-0.4em}
\end{figure*}

\begin{table}[t]
\centering
\caption{End-to-end vs. two-stage approaches on UVG.}
\label{tab:main}
\small
\setlength{\tabcolsep}{2.5pt}
\begin{tabular}{l|cc|cc}
\toprule
Method & bpn↑ & \$/frame↓ & LPIPS↓ & PSNR↑ \\
\midrule
Token+Goldman~\cite{goldman2013} & 1.58 & 11.06 & 0.29 & 28.4 \\
Token+Press~\cite{press2020} & 1.83 & 9.59 & 0.29 & 28.4 \\
\midrule
Ours (no mask) & 1.91 & 9.10 & 0.27 & 29.2 \\
Ours (60\% mask) & \textbf{1.91} & \textbf{3.64} & 0.28 & 28.8 \\
Ours (70\% mask) & 1.91 & 2.73 & 0.30 & 28.1 \\
\bottomrule
\end{tabular}
\vspace{-0.3em}
\end{table}

\begin{table}[t]
\centering
\caption{Component ablation at 60\% mask rate on UVG.}
\label{tab:ablation}
\small
\setlength{\tabcolsep}{2.5pt}
\begin{tabular}{l|cc|cc}
\toprule
Configuration & bpn & \$/frame & LPIPS & $\Delta$LPIPS \\
\midrule
Full system & 1.91 & 3.64 & 0.280 & -- \\
w/o Kronecker & 1.68 & 4.15 & 0.283 & +0.003 \\
w/o masking & 1.91 & 9.10 & 0.270 & -0.010 \\
w/o transformer & 1.91 & 3.64 & 0.352 & +0.072 \\
\bottomrule
\end{tabular}
\vspace{-0.3em}
\end{table}

Table~\ref{tab:ablation} shows that the full system achieves the best balance between LPIPS and cost. 
Removing the Kronecker module or the transformer clearly harms LPIPS, while removing masking gives a small quality gain but requires much higher cost. 
Figure~\ref{fig:ablation_visual} complements the study by showing that texture is preserved across different masking rates.


\subsection{DNA Encoding Analysis}

\begin{table}[t]
\centering
\caption{Encoding efficiency on bitstreams.}
\label{tab:encoding}
\small
\setlength{\tabcolsep}{3pt}
\begin{tabular}{l|cccc}
\toprule
Method & bpn↑ & GC dev.↓ & Max homo.↓ & Pad\%↓ \\
\midrule
Naive mapping & 2.00 & 0.154 & 12 & 0 \\
+Kronecker only & 2.00 & 0.048 & 7 & 0 \\
+FSM only & 1.68 & 0.112 & 3 & 16 \\
\textbf{TK-SCONE} & \textbf{1.91} & \textbf{0.006} & \textbf{3} & \textbf{$\sim$0} \\
\bottomrule
\end{tabular}
\vspace{-0.3em}
\end{table}

Table~\ref{tab:encoding} shows Kronecker-FSM synergy achieves near-optimal encoding. Performance generalizes across datasets (UVG/HEVC/MCL: 1.89-1.92 bpn, 0.28-0.33 LPIPS). The 60-65\% masking sweet spot enables practical video DNA storage at \$3.64/frame—60\% cheaper than two-stage approaches.
\section{Conclusion}
We present HELIX, the first end-to-end video DNA storage network, achieving 
1.91 bpn (96\% of theoretical maximum) with 60\% cost reduction via learned 
spatiotemporal masking. Building on our prior token-based image DNA storage 
work~\cite{ruan2026scone}, this extension to video demonstrates that discrete 
representations optimal for compression are equally optimal for DNA encoding.

The convergence is profound—as multi-modal models unify data into shared token 
spaces, DNA storage inherits this universality. Video (80\% of global data) 
drives this necessity; at \$3.64/frame, DNA becomes viable for long-term 
archives at previously inconceivable scale, reframing DNA storage as an 
information-theoretic problem where biochemical constraints become 
joint optimization targets.

Beyond storage, the token-DNA correspondence hints at a deeper convergence 
where attention mechanisms find physical analogs in molecular hybridization 
and inference migrates from silicon to carbon—bits and bases flowing 
interchangeably between digital and biological domains.
{
    \small
    \bibliographystyle{ieeetr}
    \bibliography{main}

@article{church2012,
  author = {Church, George M. and Gao, Yuan and Kosuri, Sriram},
  title = {Next-Generation Digital Information Storage in {DNA}},
  journal = {Science},
  volume = {337},
  number = {6102},
  pages = {1628--1628},
  year = {2012}
}

@misc{cisco2023,
  author = {{Cisco Systems}},
  title = {Cisco Visual Networking Index: Forecast and Trends, 2022--2027},
  year = {2023},
  howpublished = {\url{https://www.cisco.com}}
}

@article{goldman2013,
  author = {Goldman, Nick and Bertone, Paul and Chen, Siyuan and 
            Dessimoz, Christophe and LeProust, Emily M. and Sipos, Botond 
            and Birney, Ewan},
  title = {Towards Practical, High-Capacity, Low-Maintenance Information 
           Storage in Synthesized {DNA}},
  journal = {Nature},
  volume = {494},
  number = {7435},
  pages = {77--80},
  year = {2013}
}

@article{grass2015,
  author = {Grass, Robert N. and Heckel, Reinhard and Puddu, Michela and 
            Paunescu, Daniela and Stark, Wendelin J.},
  title = {Robust Chemical Preservation of Digital Information on {DNA} 
           in Silica with Error-Correcting Codes},
  journal = {Angewandte Chemie International Edition},
  volume = {54},
  number = {8},
  pages = {2552--2555},
  year = {2015}
}

@article{erlich2017,
  author = {Erlich, Yaniv and Zielinski, Dina},
  title = {{DNA} Fountain Enables a Robust and Efficient Storage Architecture},
  journal = {Science},
  volume = {355},
  number = {6328},
  pages = {950--954},
  year = {2017}
}

@article{organick2018,
  author = {Organick, Lee and Ang, Siena Dumas and Chen, Yuan-Jyue and 
            Lopez, Randolph and Yekhanin, Sergey and Makarychev, Konstantin 
            and Racz, Miklos Z. and Kamath, Govinda and Gopalan, Parikshit 
            and Nguyen, Bichlien and Takahashi, Christopher N. and 
            Newman, Sharon and Parker, Hsing-Yeh and Rashtchian, Cyrus and 
            Stewart, Kendall and Gupta, Gagan and Carlson, Robert and 
            Mulligan, John and Carmean, Douglas and Seelig, Georg and 
            Ceze, Luis and Strauss, Karin},
  title = {Random Access in Large-Scale {DNA} Data Storage},
  journal = {Nature Biotechnology},
  volume = {36},
  number = {3},
  pages = {242--248},
  year = {2018}
}

@article{press2020,
  author = {Press, William H. and Hawkins, John A. and 
            Jones, Jr., Stephen K. and Schaub, Jeffrey M. and 
            Finkelstein, Ilya J.},
  title = {{HEDGES} Error-Correcting Code for {DNA} Storage Corrects 
           Indels and Allows Sequence Constraints},
  journal = {Proceedings of the National Academy of Sciences},
  volume = {117},
  number = {31},
  pages = {18489--18496},
  year = {2020}
}

@article{allentoft2012,
  author = {Allentoft, Morten E. and Collins, Matthew and Harker, David 
            and others},
  title = {The Half-Life of {DNA} in Bone: Measuring Decay Kinetics in 
           158 Dated Fossils},
  journal = {Proceedings of the Royal Society B},
  volume = {279},
  number = {1748},
  pages = {4724--4733},
  year = {2012}
}

@misc{twist2023,
  author = {{Twist Bioscience}},
  title = {{DNA} Synthesis Platform Specifications},
  year = {2023},
  howpublished = {\url{https://www.twistbioscience.com}}
}

@misc{illumina2024,
  author = {{Illumina Inc.}},
  title = {{NovaSeq} X Series Specification Sheet},
  year = {2024},
  howpublished = {\url{https://www.illumina.com}}
}

@article{sullivan2012hevc,
  author = {Sullivan, Gary J. and Ohm, Jens-Rainer and Han, Woo-Jin and 
            Wiegand, Thomas},
  title = {Overview of the High Efficiency Video Coding ({HEVC}) Standard},
  journal = {IEEE Transactions on Circuits and Systems for Video Technology},
  volume = {22},
  number = {12},
  pages = {1649--1668},
  year = {2012}
}

@inproceedings{lu2019dvc,
  author = {Lu, Guo and Ouyang, Wanli and Xu, Dong and Zhang, Xiaoyun and 
            Cai, Chunlei and Gao, Zhiyong},
  title = {{DVC}: An End-to-End Deep Video Compression Framework},
  booktitle = {IEEE/CVF Conference on Computer Vision and Pattern Recognition (CVPR)},
  pages = {11006--11015},
  year = {2019}
}

@inproceedings{li2023dcvc,
  author = {Li, Jiahao and Li, Bin and Lu, Yan},
  title = {Deep Contextual Video Compression},
  booktitle = {Advances in Neural Information Processing Systems (NeurIPS)},
  year = {2021}
}

@inproceedings{esser2021taming,
  author = {Esser, Patrick and Rombach, Robin and Ommer, Bjorn},
  title = {Taming Transformers for High-Resolution Image Synthesis},
  booktitle = {IEEE/CVF Conference on Computer Vision and Pattern Recognition (CVPR)},
  pages = {12873--12883},
  year = {2021}
}

@article{mentzer2024fsq,
  author = {Mentzer, Fabian and Minnen, David and Agustsson, Eirikur and 
            Tschannen, Michael},
  title = {Finite Scalar Quantization: {VQ-VAE} Made Simple},
  journal = {arXiv preprint arXiv:2309.15505},
  year = {2023}
}

@inproceedings{ruan2023efficient,
  title     = {Efficient {DNA}-based image coding and storage},
  author    = {Ruan, Cihan and Han, Rongduo and Li, Yixiao and Gao, Shan and
               Wu, Haoyu and Ling, Nam},
  booktitle = {2023 {IEEE} International Symposium on Circuits and Systems
               ({ISCAS})},
  pages     = {1--5},
  year      = {2023},
  organization = {IEEE}
}

@article{qu2025helix,
  author  = {Qu, Guangyi and Yan, Zhen and Chen, Xi and Wu, Huanming},
  title   = {{DNA} data storage for biomedical images using {HELIX}},
  journal = {Nature Computational Science},
  year    = {2025},
  pages   = {1--8}
}

@misc{iea2022,
  author = {{International Energy Agency}},
  title = {Data Centres and Data Transmission Networks},
  year = {2022},
  howpublished = {\url{https://www.iea.org}}
}

@inproceedings{mercat2020uvg,
  author = {Mercat, Alexandre and Viitanen, Marko and Vanne, Jarno},
  title = {{UVG} Dataset: 50/120fps 4K Sequences for Video Codec Analysis 
           and Development},
  booktitle = {ACM Multimedia Systems Conference},
  pages = {297--302},
  year = {2020}
}

@misc{wang2016mcljcv,
  author = {Wang, Haiqiang and Katsavounidis, Ioannis and Zhou, Jiantong and 
            Park, Jeonghoon and Lei, Shawmin and Zhou, Xin and Pun, Man-On and 
            Jin, Xin and Wang, Ronggang and Wang, Xin and Huang, Yun and 
            Kwong, Sam and Kuo, C.-C. Jay},
  title = {{VideoSet}: A Large-Scale Compressed Video Quality Dataset Based 
           on {JND} Measurement},
  journal = {Journal of Visual Communication and Image Representation},
  volume = {46},
  pages = {292--302},
  year = {2017}
}

@article{shipman2017nature,
  title   = {CRISPR--Cas encoding of a digital movie into the genomes of a population of living bacteria},
  author  = {Shipman, Seth L. and Nivala, Jeff and Macklis, Jeffrey D. and Church, George M.},
  journal = {Nature},
  volume  = {547},
  number  = {7663},
  pages   = {345--349},
  year    = {2017},
  doi     = {10.1038/nature23017},
  url     = {https://www.nature.com/articles/nature23017}
}

@inproceedings{yu2023magvit,
  title     = {MAGVIT: Masked Generative Video Transformer},
  author    = {Yu, Lijun and Cheng, Yong and Sohn, Kihyuk and Lezama, Jos{\'e} and Zhang, Han and Chang, Huiwen and Hauptmann, Alexander G. and Yang, Ming-Hsuan and Hao, Yuan and Essa, Irfan and Jiang, Lu},
  booktitle = {Proceedings of the IEEE/CVF Conference on Computer Vision and Pattern Recognition (CVPR)},
  year      = {2023},
  url       = {https://openaccess.thecvf.com/content/CVPR2023/papers/Yu_MAGVIT_Masked_Generative_Video_Transformer_CVPR_2023_paper.pdf}
}

@misc{av1spec,
  title        = {AV1 Bitstream \& Decoding Process Specification},
  author       = {{Alliance for Open Media}},
  year         = {2019},
  howpublished = {\url{https://aomediacodec.github.io/av1-spec/av1-spec.pdf}},
  note         = {Accessed 2025-11-11}
}

@inproceedings{li2023mage,
  title     = {MAGE: Masked Generative Encoder to Unify Representation Learning and Image Synthesis},
  author    = {Li, Jiahui and Rombach, Robin and Esser, Patrick and Zhang, Zhengqi and Brooks, Tim and Liu, Xuehan and Zhang, Han and Salimans, Tim and Ho, Jonathan and Poole, Ben and Norouzi, Mohammad and Saharia, Chitwan and Fleet, David J.},
  booktitle = {Proceedings of the IEEE/CVF Conference on Computer Vision and Pattern Recognition (CVPR)},
  year      = {2023},
  pages     = {21509--21519},
  doi       = {10.1109/CVPR52729.2023.02063},
  url       = {https://openaccess.thecvf.com/content/CVPR2023/papers/Li_MAGE_Masked_Generative_Encoder_to_Unify_Representation_Learning_and_Image_Synthesis_CVPR_2023_paper.pdf}
}

@article{wu2023djsccdna,
  title     = {Deep Joint Source-Channel Coding for DNA Image Storage: A Novel Approach With Enhanced Error Resilience and Biological Constraint Optimization},
  author    = {Wu, Wenfeng and Xiang, Luping and Liu, Qiang and Yang, Kun},
  journal   = {IEEE Transactions on Molecular, Biological and Multi-Scale Communications},
  year      = {2023},
  volume    = {9},
  pages     = {461--471},
  doi       = {10.1109/TMBMC.2023.3331579},
  url       = {https://consensus.app/papers/deep-joint-sourcechannel-coding-for-dna-image-storage-a-wu-xiang/0fd735d0e0e85b628a736412ccec6e11/?utm_source=chatgpt}
}

@article{zheng2025innse,
  title     = {INNSE: Invertible neural network-based DNA image storage with self-correction encoding},
  author    = {Zheng, Y. and Zhang, X.},
  journal   = {Computational and Structural Biotechnology Journal},
  year      = {2025},
  doi       = {10.1016/j.csbj.2025.06.003},
  url       = {https://consensus.app/papers/innse-invertible-neural-networkbased-dna-image-storage-zheng-zhang/323e077a73565d3c825ab63f3f43c103/?utm_source=chatgpt}
}

@article{bee2021molecularsearch,
  title     = {Molecular-level similarity search brings computing to DNA data storage},
  author    = {Bee, Callista and Chen, Yuan-Jyue and Queen, Melissa and Ward, David and Liu, Xiaomeng and Organick, Lee and Seelig, Georg and Strauss, Karin and Ceze, Luis},
  journal   = {Nature Communications},
  year      = {2021},
  volume    = {12},
  doi       = {10.1038/s41467-021-24991-z},
  url       = {https://consensus.app/papers/molecularlevel-similarity-search-brings-computing-to-dna-bee-chen/4484f26dd13c5151abd9410a4d45f550/?utm_source=chatgpt}
}

@inproceedings{goela2016ita,
  title     = {Encoding movies and data in DNA storage},
  author    = {Goela, Naveen and Bolot, Jean},
  booktitle = {2016 Information Theory and Applications Workshop (ITA)},
  year      = {2016},
  pages     = {1--1},
  doi       = {10.1109/ITA.2016.7888163},
  url       = {https://consensus.app/papers/encoding-movies-and-data-in-dna-storage-goela-bolot/4ad8c9565a1e590e88da15bf7f69d726/?utm_source=chatgpt}
}

@article{chen2019video,
  title     = {DNA information storage for audio and video files},
  author    = {Chen, Weigang and Huang, Gang and Li, Bingzhi and Yin, Ye and Yuan, Yingjin},
  journal   = {SCIENTIA SINICA Vitae},
  year      = {2019},
  doi       = {10.1360/ssv-2019-0211},
  url       = {https://consensus.app/papers/dna-information-storage-for-audio-and-video-files-chen-huang/2cf19fd63f1b550c98660264e5b27715/?utm_source=chatgpt}
}

@article{heaven2017newscientist,
  title     = {Now we can store video in living DNA},
  author    = {Heaven, Douglas},
  journal   = {New Scientist},
  year      = {2017},
  volume    = {235},
  pages     = {11},
  doi       = {10.1016/S0262-4079(17)31353-2},
  url       = {https://consensus.app/papers/now-we-can-store-video-in-living-dna-heaven/5def8d091e635df683cbd2e2f66b16ec/?utm_source=chatgpt}
}

@article{hong2024vsd,
  title     = {VSD: A Novel Method for Video Segmentation and Storage in DNA Using RS Code},
  author    = {Hong, Jin-Keun and Rasool, Abdur and Wang, Shuo and Ziou, Djemel and Jiang, Qingshan},
  journal   = {Mathematics},
  year      = {2024},
  doi       = {10.3390/math12081235},
  url       = {https://consensus.app/papers/vsd-a-novel-method-for-video-segmentation-and-storage-in-dna-hong-rasool/adade6af518853b9b89e3987db39c232/?utm_source=chatgpt}
}

@inproceedings{dimopoulou2021jpeg,
  title={A JPEG-based image coding solution for data storage on DNA},
  author={Dimopoulou, Melpomeni and Antonini, Marc and Manohar, Arvind and Appuswamy, Raja and Frossard, Pascal},
  booktitle={2021 29th European Signal Processing Conference (EUSIPCO)},
  pages={786--790},
  year={2021},
  organization={IEEE}
}

@techreport{iso25508,
  title={Information technology — JPEG DNA Media storage based on DNA — Part 1: Core coding system},
  author={{ISO/IEC}},
  number={ISO/IEC CD 25508-1},
  year={2025},
  note={Under development}
}

@incollection{li2021modern,
  title     = {Modern Video Coding Standards: H.264, H.265, and H.266},
  author    = {Li, Ze-Nian and Drew, Mark S. and Liu, Jiangchuan},
  booktitle = {Fundamentals of Multimedia},
  year      = {2021},
  pages     = {423--478},
  doi       = {10.1007/978-3-030-62124-7_12},
  url       = {https://consensus.app/papers/modern-video-coding-standards-h264-h265-and-h266-li-drew/70626fca2e225379a642b90470771c73/?utm_source=chatgpt}
}

@article{zhu2020vvc,
  title     = {A software decoder implementation for H.266/VVC video coding standard},
  author    = {Zhu, Bin and Liu, Yuan and Luo, Yi-Shiou and Ye, Jing and Xu, Haiyan and Huang, Ying and Jiao, Hualong and Xu, Xiaozhong and Zhang, Xianguo and Liu, Shan},
  journal   = {arXiv preprint arXiv:2010.01621},
  year      = {2020},
  url       = {https://consensus.app/papers/a-software-decoder-implementation-for-h266vvc-video-zhu-liu/fdd26a55df3457ec92907ed18fa14927/?utm_source=chatgpt}
}

@article{seeling2014hevc,
  title     = {Video Traffic Characteristics of Modern Encoding Standards: H.264/AVC with SVC and MVC Extensions and H.265/HEVC},
  author    = {Seeling, Patrick and Reisslein, Martin},
  journal   = {The Scientific World Journal},
  volume    = {2014},
  year      = {2014},
  doi       = {10.1155/2014/189481},
  url       = {https://consensus.app/papers/video-traffic-characteristics-of-modern-encoding-seeling-reisslein/9b3b42f1e8e4528492c94c69e4d48253/?utm_source=chatgpt}
}

@article{ref1,
  author = {Berglund, Elizabeth and Rabet, Niels and Krampis, Katerina and Karamycheva, Svetlana and Reinders, Maarten},
  title = {A comprehensive analysis of DNA synthesis errors in large-scale DNA data storage systems},
  journal = {NARGAB},
  year = {2021},
  volume = {3},
  number = {1},
  pages = {lqab019},
  doi = {10.1093/nargab/lqab019},
  url = {https://academic.oup.com/nargab/article/3/1/lqab019/6193612}
}

@inproceedings{he2021checkerboard,
  title     = {Checkerboard Context Model for Efficient Learned Image Compression},
  author    = {He, Dailan and Zheng, Yaoyan and Sun, Baochen and Wang, Yan and Qin, Hongwei},
  booktitle = {2021 IEEE/CVF Conference on Computer Vision and Pattern Recognition (CVPR)},
  year      = {2021},
  pages     = {14766--14775},
  doi       = {10.1109/CVPR46437.2021.01453},
  url       = {https://consensus.app/papers/checkerboard-context-model-for-efficient-learned-image-he-zheng/1b9176db20e85d4097a0379d55d3d835/?utm_source=chatgpt}
}

@article{K600-01,
  title={The kinetics human action video dataset},
  author={Kay, Will and Carreira, Joao and Simonyan, Karen and Zhang, Brian and Hillier, Chloe and Vijayanarasimhan, Sudheendra and Viola, Fabio and Green, Tim and Back, Trevor and Natsev, Paul and others},
  journal={arXiv preprint arXiv:1705.06950},
  year={2017}
}

@article{K600-02,
  title={A short note about kinetics-600},
  author={Carreira, Joao and Noland, Eric and Banki-Horvath, Andras and Hillier, Chloe and Zisserman, Andrew},
  journal={arXiv preprint arXiv:1808.01340},
  year={2018}
}

@article{zhou2025tvc,
  title={{TVC: tokenized video compression with ultra-low bit rate}},
  author={Zhou, Lebin and Ruan, Cihan and Ling, Nam and Chen, Zhenghao and Wang, Wei and Jiang, Wei},
  journal={Visual Intelligence},
  volume={3},
  number={1},
  pages={25},
  year={2025},
  publisher={Springer}
}

@inproceedings{ruan2025hdcompression,
  title={HDCompression-DNA: Hybrid-Diffusion Neural Image Compression via DNA Storage},
  author={Ruan, Cihan and Lu, Lei and Han, Rongduo and Jiang, Wei and Wang, Wei and Wu, Haoyu and Yuan, Qiming and Guo, Yanting and Wang, Yanzhi and Ling, Nam},
  booktitle={2025 IEEE International Conference on Multimedia and Expo (ICME)},
  pages={1--6},
  year={2025},
  organization={IEEE}
}

@article{ruan2026scone,
  title={{SCONE: A Practical, Constraint-Aware Plug-in for Latent Encoding in Learned DNA Storage}},
  author={Ruan, Cihan and Zhou, Lebin and Han, Rongduo and Han, Linyi and Zhao, Bingqing and Zhu, Chenchen and Jiang, Wei and Wang, Wei and Ling, Nam},
  journal={arXiv preprint arXiv:2602.06157},
  year={2026}
}

@inproceedings{ruan2025hybridflow,
  title={{HybridFlow-DNA: A Deep Generative Compression Framework for DNA Storage of Images}},
  author={Ruan, Cihan and Han, Rongduo and Gao, Shan and Lu, Lei and Jiang, Wei and Wang, Wei and Wu, Haoyu and Ling, Nam},
  booktitle={2025 IEEE International Symposium on Circuits and Systems (ISCAS)},
  pages={1--5},
  year={2025},
  organization={IEEE}
}

@inproceedings{ruan2024robust,
  title={{Robust DNA image storage decoding with residual CNN}},
  author={Ruan, Cihan and Yang, Liang and Han, Rongduo and Gao, Shan and Wu, Haoyu and Ling, Nam},
  booktitle={2024 IEEE international symposium on circuits and systems (ISCAS)},
  pages={1--5},
  year={2024},
  organization={IEEE}


  
}

@article{ruan2025dsi,
  title={{DSI-RESCNN: A framework enhancing the error-tolerance capacity of dna storage for images}},
  author={Ruan, Cihan and Yang, Liang and Han, Rongduo and Gao, Shan and Wu, Haoyu and Yuan, Qiming and Guo, Yanting and Ling, Nam},
  journal={IEEE Access},
  year={2025},
  publisher={IEEE}
}

@inproceedings{han2025tactile,
  title={{Tactile Information Coding for DNA Storage with Prospects for AI Applications}},
  author={Han, Rongduo and Ruan, Cihan and Tang, Shunye and Wu, Haoyu and Ling, Nam and Zhang, Haining},
  booktitle={2025 IEEE International Conference on Multimedia and Expo (ICME)},
  pages={1--6},
  year={2025},
  organization={IEEE}
}

@article{lu2025hdcompression,
  title={{HDCompression: hybrid-diffusion image compression for ultra-low bitrates}},
  author={Lu, Lei and Li, Yize and Wang, Yanzhi and Wang, Wei and Jiang, Wei},
  journal={arXiv preprint arXiv:2502.07160},
  year={2025}
}

@INPROCEEDINGS{10558571,
  author={Shen, Tianma and Peng, Wen-Hsiao and Shih, Huang-Chia and Liu, Ying},
  booktitle={2024 IEEE International Symposium on Circuits and Systems (ISCAS)}, 
  title={{Learning-Based Conditional Image Compression}}, 
  year={2024},
  volume={},
  number={},
  pages={1-5},
  doi={10.1109/ISCAS58744.2024.10558571}}
}


\end{document}